\documentclass[sigconf]{acmart}

\usepackage{amsthm}
\usepackage{multirow}
\usepackage{stfloats}
\usepackage{graphicx}
\usepackage{url}
\usepackage{subcaption}
\usepackage[disable]{todonotes}

\AtBeginDocument{%
  }

\copyrightyear{2024}
\acmYear{2024}
\setcopyright{acmlicensed}
\acmConference[KDD '24]{Proceedings of the 30th
ACM SIGKDD Conference on Knowledge Discovery and Data Mining}{August
25--29, 2024}{Barcelona, Spain}
\acmBooktitle{Proceedings of the 30th ACM SIGKDD Conference on Knowledge
Discovery and Data Mining (KDD '24), August 25--29, 2024, Barcelona, Spain}
\acmDOI{10.1145/3637528.3671961}
\acmISBN{979-8-4007-0490-1/24/08}

\begin{document}

\title{Heterogeneity-Informed Meta-Parameter Learning for Spatiotemporal Time Series Forecasting}

\settopmatter{authorsperrow=4}


\author{Zheng Dong}
\authornote{Equal contribution.}
\affiliation{
  \institution{Southern University of Science and Technology}
  \city{Shenzhen}
  \country{China}
}
\email{zhengdong00@outlook.com}

\author{Renhe Jiang}
\authornotemark[1]
\affiliation{
  \institution{The University of Tokyo}
  \city{Tokyo}
  \country{Japan}
}
\email{jiangrh@csis.u-tokyo.ac.jp}

\author{Haotian Gao}
\affiliation{
  \institution{The University of Tokyo}
  \city{Tokyo}
  \country{Japan}
}
\email{gaoht6@outlook.com}

\author{Hangchen Liu}
\affiliation{
  \institution{Southern University of Science and Technology}
  \city{Shenzhen}
  \country{China}
}
\email{liuhc3@outlook.com}

\author{Jinliang Deng}
\affiliation{
  \institution{Hong Kong University of Science and Technology}
  \city{Hong Kong}
  \country{China}
}
\email{dengjinliang@ust.hk}

\author{Qingsong Wen}
\affiliation{
  \institution{Squirrel AI}
  \city{Seattle}
  \country{USA}
}
\email{qingsongedu@gmail.com}

\author{Xuan Song}
\authornote{Corresponding author.}
\affiliation{
  \institution{Jilin University}
  \city{Changchun}
  \country{China}
}
\affiliation{
    \institution{Southern University of Science and Technology}
    \city{Shenzhen}
    \country{China}
}
\email{songx@sustech.edu.cn}

\renewcommand{\shortauthors}{Zheng Dong et al.}



\begin{abstract}
Spatiotemporal time series forecasting plays a key role in a wide range of real-world applications. While significant progress has been made in this area, fully capturing and leveraging spatiotemporal heterogeneity remains a fundamental challenge. Therefore, we propose a novel \textbf{Heterogeneity-Informed Meta-Parameter Learning} scheme. Specifically, our approach implicitly captures spatiotemporal heterogeneity through learning spatial and temporal embeddings, which can be viewed as a clustering process. Then, a novel spatiotemporal meta-parameter learning paradigm is proposed to learn spatiotemporal-specific parameters from meta-parameter pools, which is informed by the captured heterogeneity. Based on these ideas, we develop a \underline{H}eterogeneity-\underline{I}nformed Spatiotemporal \underline{M}eta-\underline{Net}work (\textbf{HimNet}) for spatiotemporal time series forecasting. Extensive experiments on five widely-used benchmarks demonstrate our method achieves state-of-the-art performance while exhibiting superior interpretability. Our code is available at \underline{\url{https://github.com/XDZhelheim/HimNet}}.
\end{abstract}


\maketitle

\begin{figure}[!t]
    \centering
    \includegraphics[width=\linewidth]{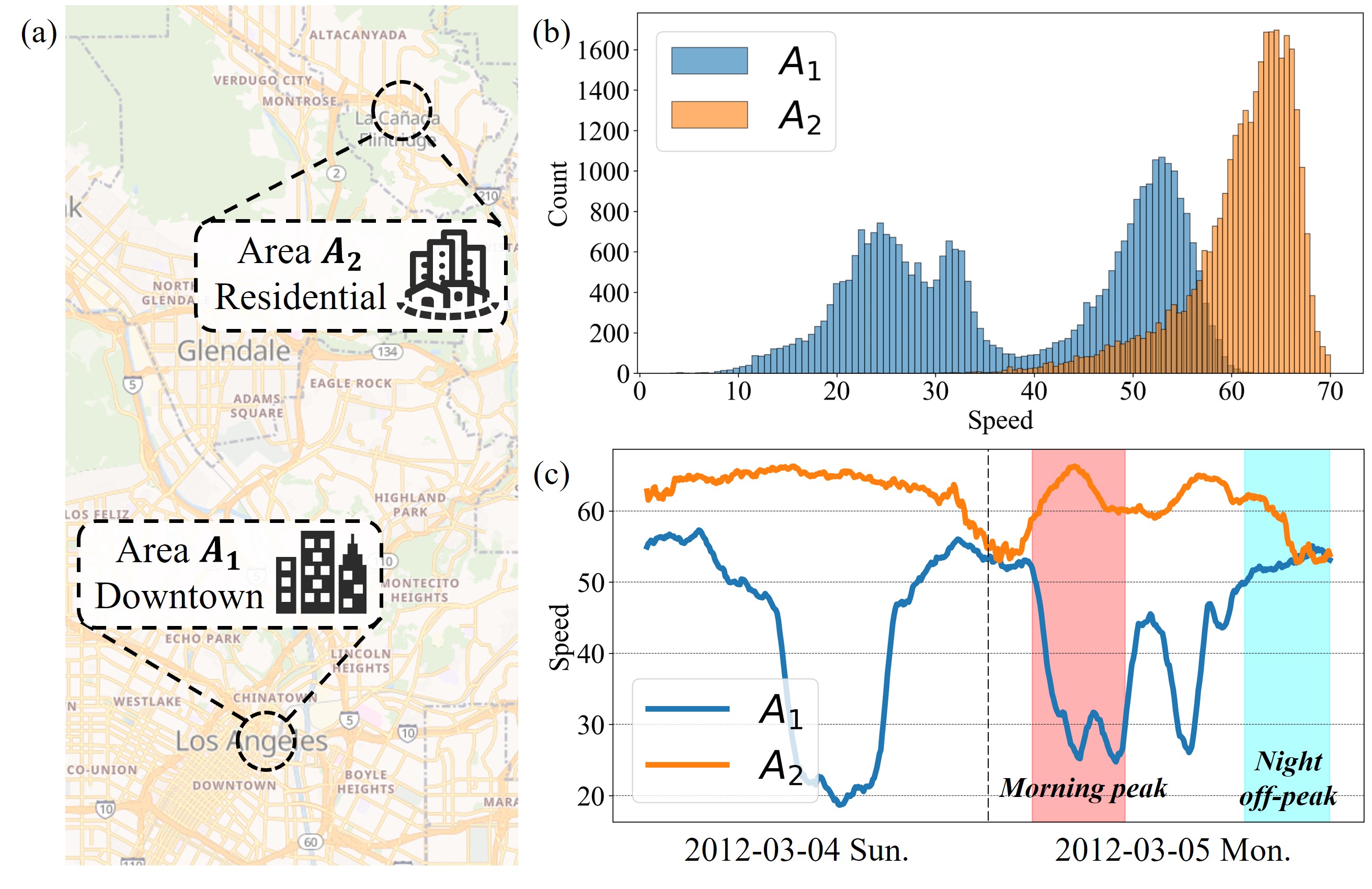}
    \caption{Spatiotemporal heterogeneity. (a): Locations of two selected areas. $A_1$ is in the downtown region, while $A_2$ is a residential area. (b): Illustration of spatial heterogeneity, with differing vehicle speed distributions between $A_1$ and $A_2$. (c): Demonstration of temporal heterogeneity, showing distinct patterns between different time periods.}
    \label{fig: intro}
\end{figure}

\section{Introduction}

The rise of sensor networks has brought the widespread collection of spatiotemporal time series in urban environments. As these datasets grow quickly, accurate spatiotemporal time series forecasting is increasingly important for many real-world applications such as transportation~\cite{DLTraff,TransGTR}, weather~\cite{weather}, and economics~\cite{LSTNet}. While considerable efforts~\cite{DMSTGCN,STGODE,ExploringProgressInMTS,STWave} have been made, existing methods still face a major challenge: fully capturing and leveraging \textbf{\textit{spatiotemporal heterogeneity}}. Spatial heterogeneity refers to the phenomenon that distinct patterns are observed across different location types during the same time period, while temporal heterogeneity is reflected by the unique patterns observed at the same location across different time periods. Figure~\ref{fig: intro} provides a real-world example. As shown in Figure~\ref{fig: intro}(a), we analyze two areas in different spatial contexts: Area $A_1$ in downtown LA and Area $A_2$ in a residential region. Figure~\ref{fig: intro}(b) reveals substantial differences in vehicle speed distributions between them: $A_1$ shows a lower average speed due to daily downtown congestion, while $A_2$ exhibits higher speeds with less commuting pressure. This demonstrates spatial heterogeneity. Figure~\ref{fig: intro}(c) illustrates their time series in different temporal contexts. The patterns are stable during the holiday (Sun.) but fluctuate violently during the workday (Mon.). Furthermore, morning peak hours and nighttime off-peak hours also exhibit opposing trends, showing temporal heterogeneity. These phenomena occur consistently throughout spatiotemporal data. Thus, effectively modeling spatiotemporal heterogeneity is critical to accurate forecasting. As time series from diverse spatial locations and time periods display distinct characteristics, proper capturing of heterogeneity relies on identifying and distinguishing such varied spatiotemporal contexts.

To account for different contexts, graph-based approaches~\cite{STSGCN,STFGNN,SHGNN} attempt to capture heterogeneity by utilizing handcrafted features such as graph topology and similarity metrics between time series. However, depending heavily on \textbf{pre-defined representations} limits their adaptability and generalizability, as these features cannot encompass the full complexity of diverse contexts. To enhance adaptability, other works introduce meta-learning techniques~\cite{STMetaNet,STMetaNet+} that apply multiple parameter sets across different spatial locations. However, they also rely on \textbf{auxiliary characteristics} including Points-of-Interests (POIs) and sensor placements. Additionally, the \textbf{high computational and memory costs} limit their applicability to large-scale datasets. Recent representation learning methods~\cite{STID,STAEformer} effectively identify heterogeneity through input embeddings. However, their oversimplified downstream processing structures \textbf{fail to fully leverage} the representational power. While self-supervised learning methods~\cite{ST-SSL,STDMAE} also succeed in capturing spatiotemporal heterogeneity by designing additional tasks, fully \textbf{end-to-end} joint optimization for spatiotemporal forecasting continues to present difficulties for these approaches.
In short, the key drawbacks are: (1) reliance on auxiliary features; (2) high computational and memory costs; (3) failure to fully leverage captured heterogeneity; and (4) difficulties with end-to-end optimization.

To address the above limitations in prior works, we propose a novel \textbf{Heterogeneity-Informed Meta-Parameter Learning} scheme and a \underline{H}eterogeneity-\underline{I}nformed \underline{M}eta-\underline{Net}work (\textbf{HimNet}) for spatiotemporal time series forecasting. In detail, we first implicitly characterize heterogeneity by learning spatial and temporal embeddings from a clustering view. The representations gradually differentiate and form distinct clusters during training, capturing underlying spatiotemporal contexts. Next, a novel spatiotemporal meta-parameter learning paradigm is proposed to enhance model adaptability and generalizability, which is flexible for various domains. Specifically, we learn a unique parameter set for each spatiotemporal context via querying a small meta-parameter pool. Based on these, we further propose Heterogeneity-Informed Meta-Parameter Learning that uses the characterized heterogeneity to inform meta-parameter learning. Thus, our approach can not only capture but explicitly leverage spatiotemporal heterogeneity to improve forecasting. Finally, we design an end-to-end network called HimNet, implementing these techniques for spatiotemporal forecasting. In summary, our contributions are three-fold:
\begin{itemize}
    \item Methodologically, we present a novel HimNet model for spatiotemporal time series forecasting. It captures inherent spatiotemporal heterogeneity through learnable embeddings, which then inform spatiotemporal meta-parameter learning to enhance model adaptability and generalizability. 
    \item Theoretically, to the best of our knowledge, our proposed Heterogeneity-Informed Meta-Parameter Learning is the first method that not only captures but also fully leverages spatiotemporal heterogeneity. This enables our model to distinguish and adapt to different spatiotemporal contexts.
    \item Empirically, HimNet significantly outperforms state-of-the-art methods on five benchmarks based on extensive experiments, demonstrating superior performance with competitive efficiency. Visualization experiments further illustrate its strong interpretability through meta-parameters.
\end{itemize}

\section{Related Work}

\subsection{Spatiotemporal Forecasting}
Spatiotemporal forecasting has seen extensive research as it plays a key role in many real-world applications~\cite{LargeST,MemeSTN,SpatiotemporalDiffusionPointProcesses,liuhao2023,Greto,CDSTG,GMRL,MoSSL,zhou2020foresee}. Early approaches relied on traditional time series analysis methods such as ARIMA~\cite{ARIMA-traffic} and VAR~\cite{VAR}, as well as machine learning techniques including $k$-NN~\cite{kNN-traffic} and SVM~\cite{SVM-traffic}, but they often fail to capture complex spatiotemporal dependencies inherent to the data. Recent years have witnessed remarkable progress in deep learning methods. Recurrent Neural Networks (RNNs) like LSTMs~\cite{LSTM-traffic-2015,LSTM-traffic-2018} and GRUs~\cite{GRU} achieved performance gains by effectively modeling the temporal dynamics. Convolutional networks such as WaveNet~\cite{WaveNet} also found success via their long receptive fields. However, these models do not fully represent spatial dependencies critical to networked urban systems. To address this limitation, Graph Convolutional Networks (GCNs) have been extensively explored for spatiotemporal forecasting. Pioneering works such as STGCN~\cite{STGCN} and DCRNN~\cite{DCRNN} achieved better performance by integrating GCNs with temporal models~\cite{GWNet,AGCRN,StemGNN,MTGNN}. Building on this foundation, many innovative methods have been proposed in recent years~\cite{GTS,MVMT,DSTAGNN,PMMemNet,MegaCRN,SCNN,CaST,jin2023survey}. Moreover, Transformer~\cite{attention} has also revolutionized the field, motivating time series transformers~\cite{TFTimeSeriesSurvey,chen2024multi} that expertly capture spatiotemporal correlations~\cite{GMAN,PDFormer,STAEformer,TESTAM} or handle long sequences~\cite{Informer,Autoformer,FEDformer,iTransformer}. While showing impressive performance, existing work still lacks further exploration of spatiotemporal heterogeneity. Our study aims to fill this gap through a novel HimNet that captures and leverages the heterogeneity for improved spatiotemporal forecasting.

\subsection{Meta-Parameter Learning}
One of the earliest works~\cite{Fast-weight} proposed predicting network parameters for temporal modeling. More recently, Hypernetworks~\cite{HyperNetworks} applied this idea to recurrent networks by generating adaptive weights, acting as a form of weight-sharing across layers. Learnet~\cite{learnet} was constructed as a second deep network to predict the parameters of another deep model. Dynamic Filter Networks~\cite{DFN} dynamically generate convolutional filters conditioned on input, increasing flexibility due to their adaptive nature. For meta multi-task learning, MetaMTL~\cite{MetaMTL} introduced a shared Meta-LSTM to generate basic LSTM parameters based on the current input context. In spatiotemporal applications, ST-GFSL~\cite{ST-GFSL} proposed learning non-shared parameters for cross-city transfer using learned spatiotemporal meta-knowledge from the input city. While effective in other domains, these methods were not designed specifically for spatiotemporal forecasting. To address this, recent works have introduced meta-parameter learning techniques for spatiotemporal data. ST-MetaNet~\cite{STMetaNet} first applied meta-learning across spatial locations, using auxiliary POI and road network information as spatial meta-knowledge to learn multiple parameter sets. AGCRN~\cite{AGCRN} proposed node adaptive parameter learning to generate node-specific parameters from a learned embedding matrix, which can be interpreted as learning node specific patterns from a set of candidate patterns discovered from all spatiotemporal time series. ST-WA~\cite{STWA} proposes a spatiotemporal aware approach to jointly learn location-specific and time-varying model parameters from encoded stochastic variables. In contrast to these approaches, as shown in Table~\ref{table: related_work}, HimNet learns meta-parameters using only the spatiotemporal time series itself, without depending on auxiliary features. Critically, we propose a concurrent learning scheme that operates on the temporal, spatial, and spatiotemporal joint dimensions simultaneously. Therefore, HimNet achieves maximum adaptability to any kind of spatiotemporal context. This makes our approach highly flexible and capable of extracting the full informational value from learned spatiotemporal heterogeneity.

\begin{table}[h]
\small
\centering
\caption{Comparison of related meta-parameter learning methods for spatiotemporal forecasting.}
\label{table: related_work}
\renewcommand\arraystretch{1.1}
\begin{tabular}{l|c|ccc}
\toprule
Method & Data-Independent & Temporal & Spatial & Joint \\
\hline
ST-MetaNet~\cite{STMetaNet} & $\times$ & $\times$ & \checkmark & $\times$ \\
AGCRN~\cite{AGCRN} & \checkmark & $\times$ & \checkmark & $\times$ \\
ST-WA~\cite{STWA} & \checkmark & $\times$ & $\times$ & \checkmark \\
\textbf{HimNet} & \checkmark & \checkmark & \checkmark & \checkmark \\
\bottomrule
\end{tabular}%
\end{table}

\section{Problem Definition}
Given a spatiotemporal time series $X_{t-(T-1):t}$ over the past $T$ time steps, our goal is to forecast the values over the future $T'$ time steps. That is, we aim to map $[X_{t-(T-1)}, ..., X_{t}]$ $\rightarrow$ $[X_{t+1}, ..., X_{t+T'}]$, where each $X_i$ $\in$ $\mathbb{R}^{N}$ represents the observations at the $i$-th time step for $N$ time series, typically from sensors deployed at $N$ locations.


\begin{figure*}[!t]
    \centering
    \includegraphics[width=\linewidth]{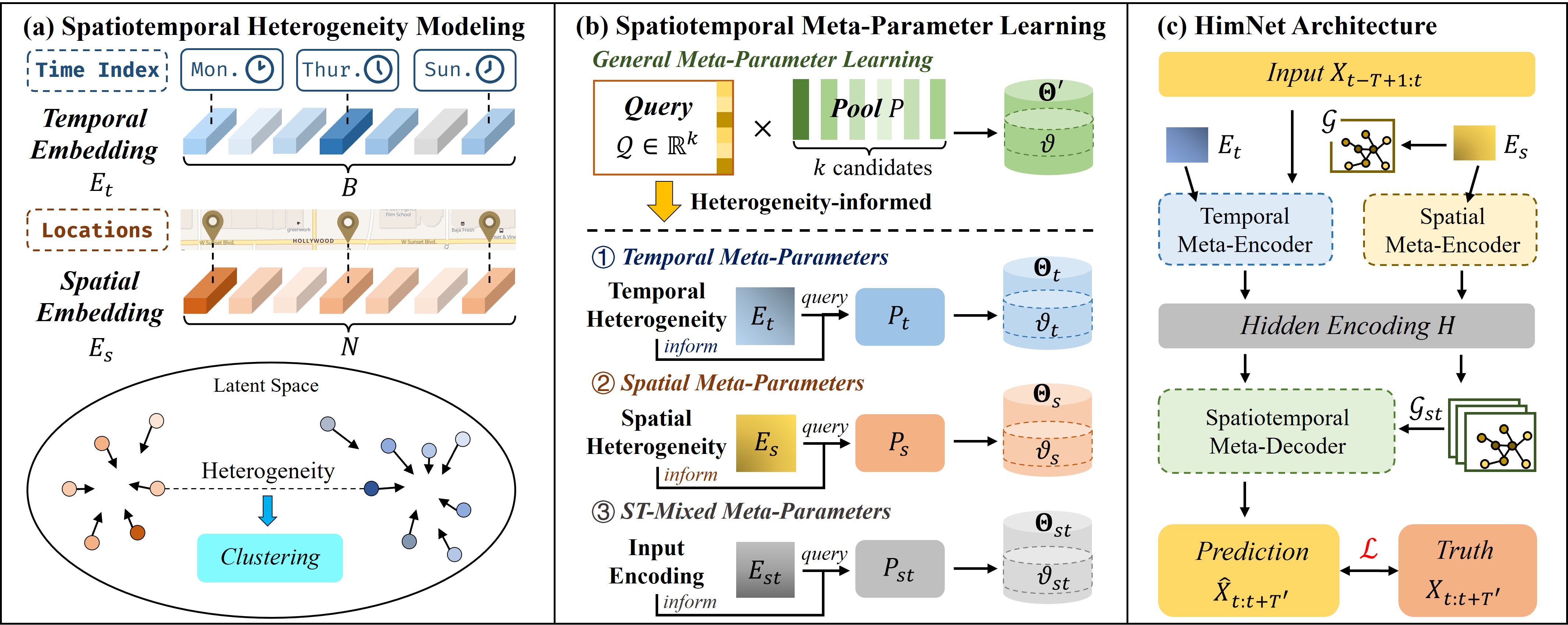}
    \caption{The overall illustration of the proposed Heterogeneity-Informed Meta-Parameter Learning and HimNet model.}
    \label{fig: method}
\end{figure*}

\section{Methodology}
In this section, we present details of our proposed Heterogeneity-Informed Meta-parameter Learning scheme along with the HimNet model, where the key components are depicted in Figure~\ref{fig: method}. To better illustrate our proposed method, we consider a minibatch dimension $B$ in the following discussions. For ease of understanding, $B$ can be assumed to be 1 without loss of generality.

\subsection{Spatiotemporal Heterogeneity Modeling}

A key aspect of modeling spatiotemporal heterogeneity is identifying and distinguishing input contexts across both temporal and spatial dimensions~\cite{STNorm}. Rather than relying on auxiliary data~\cite{STMetaNet,STSGCN}, we employ learnable embeddings~\cite{STID,STAEformer}, which assign a unique representation to each spatiotemporal context.

For the temporal dimension, we construct a time-of-day embedding dictionary $D_{tod}\in\mathbb{R}^{N_d\times d_{tod}}$ and a day-of-week dictionary $D_{dow}\in\mathbb{R}^{N_w\times d_{dow}}$, where $d_{tod}$ and $d_{dow}$ are the embedding dimensions, $N_d$ denotes the number of timesteps per day, and $N_w$ represents the number of days in a week. Given a minibatch of input samples $X^b\in\mathbb{R}^{B\times T\times N}$ with historical length $T$, the last step's timestamp of each sample in the minibatch is used as the time index to extract the corresponding time-of-day embedding $E_{tod}\in\mathbb{R}^{B\times d_{tod}}$ and day-of-week embedding $E_{dow}\in\mathbb{R}^{B\times d_{dow}}$ from the dictionaries. The two embeddings are concatenated to obtain the overall temporal embedding $E_t\in\mathbb{R}^{B\times d_{t}}$:
\begin{equation}
    E_t=E_{tod}||E_{dow},
\end{equation}
where $d_t=d_{tod}+d_{dow}$ represents the temporal embedding dimension and $||$ denotes concatenation operation. Time-of-day identifies periodic patterns with fine-grained time intervals, capturing phenomena like peak and off-peak hours. Day-of-week identifies meaningful longer-term patterns, such as the distinction in time series between weekdays and weekends. Leveraging these dictionaries, the temporal embedding aims to learn representations that can implicitly recognize and account for such temporal heterogeneity over multiple timescales.

For the spatial dimension, we utilize a spatial embedding matrix $E_s\in\mathbb{R}^{N\times d_s}$, where $N$ denotes the total number of time series or spatial locations in the dataset (e.g. number of sensors) and $d_s$ is the dimension of the embedding vector assigned to each location. Unlike the temporal dimension where the embeddings are extracted from learned dictionaries, here each of the $N$ spatial locations is directly associated with a unique learnable spatial embedding vector initialized randomly in the embedding matrix. The goal of introducing $E_s$ is to account for the inherent functional differences between locations that can influence time series patterns, such as factors related to infrastructure and urban design. It aims to learn latent representations that can help distinguish such diverse contexts, without the need for auxiliary geographical data. This facilitates modeling the spatial heterogeneity present in each location.

Furthermore, learning these embedding vectors essentially performs a dynamic clustering process. During model training, the representations within the embedding matrices gradually differentiate based on the input time series. Embedding vectors belonging to spatiotemporal contexts that exhibit similar trends in their time series will move closer in the latent space, and vice versa. As a result, this dynamic adjustment of the embeddings can be viewed as a clustering process, where representations organically form clusters. Each distinct cluster captures a typical context in the spatiotemporal time series. Analogous to the NLP models where the learned embeddings of "man" and "woman" can naturally be very close, this clustering process could be successfully achieved without involving any additional constraints like the way did in~\cite{MegaCRN,ST-SSL}. This clustering perspective is what empowers the model to distinguish different contexts and thereby accurately model the heterogeneity across space and time. Through the learned embeddings, spatiotemporal heterogeneity is modeled in an intrinsic, data-driven manner, without requiring manual feature engineering or domain expertise to describe it in advance.

\subsection{Spatiotemporal Meta-Parameter Learning}


To fully leverage the heterogeneity encoded in the spatiotemporal embeddings, we propose Heterogeneity-Informed Meta-Parameter Learning. It provides a simple yet effective way to improve model adaptability and generalizability. 

We begin by introducing the general meta-parameter learning paradigm. For the model's original parameter space $\Theta$ with size $S$, our approach first extends it along a specific dimension to create an enlarged parameter space $\Theta'$. Specifically, we can build three enlarged spaces for the temporal and spatial dimensions: temporal parameter space $\Theta_t\in\mathbb{R}^{\mathcal T\times S}$, spatial parameter space $\Theta_s\in\mathbb{R}^{N\times S}$, and spatiotemporal joint space $\Theta_{st}\in\mathbb{R}^{\mathcal T\times N\times S}$. Here, $\mathcal T$ represents the total number of timesteps and $N$ is the number of spatial locations in the dataset. This extension aims to assign a unique parameter set to each spatiotemporal context, rather than sharing $\Theta$. Directly optimizing these enlarged parameter spaces is prohibitively expensive, as they become very huge, especially when $\mathcal T$ and $N$ are large, which will lead to exploding computational and memory costs. To address this challenge, we maintain a small meta-parameter pool for each parameter space, which contains $k$ parameter candidates. The pool size $k$ is defined to be much smaller than $\mathcal T$ and $N$, but sufficiently large to encapsulate the variety of different contexts. For an incoming query representing a specific spatiotemporal context, we dynamically generate its meta-parameter as a weighted combination of the pool candidates.

For a meta-parameter pool $P\in\mathbb{R}^{k\times S}$, given a query $\mathcal Q\in\mathbb{R}^k$, the general formulation of our meta-parameter learning paradigm is:
\begin{equation}
    \vartheta=\mathcal Q\cdot P,
\label{eq: meta-param}
\end{equation}
where $\vartheta\in\mathbb{R}^S$ is the generated meta-parameter. By maintaining meta-parameter pools, our approach alleviates this computational burden by optimizing within the compact pools rather than the full enlarged parameter spaces. Thus, the complexity is reduced from $O(\mathcal {T}N)$ to $O(k)$, where $k \ll \mathcal{T}N$. This makes the learning computationally feasible for large spatiotemporal datasets.

Specifically, we leverage the embeddings proposed earlier that encode spatiotemporal heterogeneity as queries to adaptively generate the meta-parameters for each spatiotemporal context. This results in a \textit{\textbf{Heterogeneity-Informed Meta-Parameter Learning}} approach. Based on Equation~\ref{eq: meta-param}, three types of meta-parameters are generated for the respective enlarged parameter spaces.

\noindent\textbf{Temporal Meta-Parameters.} We maintain a small temporal meta-parameter pool $P_t\in\mathbb{R}^{d_t\times S}$ containing $d_t$ candidates. Using the temporal embedding $E_t\in\mathbb{R}^{B\times d_{t}}$ as queries, the temporal meta-parameters $\vartheta_t\in\mathbb{R}^{B\times S}$ are generated by:
\begin{equation}
    \vartheta_t=E_t\cdot P_t.
\label{eq: TMP}
\end{equation}

\noindent\textbf{Spatial Meta-Parameters.} We maintain a small spatial meta-parameter pool $P_s\in\mathbb{R}^{d_s\times S}$ with $d_s$ candidates. Using the spatial embedding $E_s\in\mathbb{R}^{N\times d_{s}}$ as queries, the spatial meta-parameters $\vartheta_s\in\mathbb{R}^{N\times S}$ for each location are generated by:
\begin{equation}
    \vartheta_s=E_s\cdot P_s.
\label{eq: SMP}
\end{equation}

\noindent\textbf{ST-Mixed Meta-Parameters.} To directly create a joint spatiotemporal embedding is prohibitive, as its size would also be too large to optimize effectively. Instead, we utilize the rich information inherently contained within the input data itself~\cite{STWA}, where the time series patterns naturally reflect spatiotemporal heterogeneity. For a minibatch of input samples $X^b\in\mathbb{R}^{B\times T\times N}$, we encode it into a spatiotemporal embedding $E_{st}\in\mathbb{R}^{B\times N\times d_{st}}$ as:
\begin{equation}
    E_{st}=F_{enc}(X^b),
\end{equation}
where $d_{st}$ is the embedding dimension, and $F_{enc}(\cdot)$ represents an encoding function such as linear projection, TCN~\cite{STGCN}, or GRU~\cite{DCRNN} designed to encode the input into a hidden representation.

Accordingly, we maintain a spatiotemporal meta-parameter pool $P_{st}\in\mathbb{R}^{d_{st}\times S}$ having $d_{st}$ candidates. Leveraging the encoded spatiotemporal embedding $E_{st}$ as queries, the spatiotemporal-mixed (ST-mixed) meta-parameters $\vartheta_{st}\in\mathbb{R}^{B\times N\times S}$ are generated by:
\begin{equation}
    \vartheta_{st}=E_{st}\cdot P_{st}.
\label{eq: STMP}
\end{equation}

While we generate the meta-parameters based on the proposed embeddings, the underlying meta-parameter learning technique itself is not limited specifically to these embeddings. We present a purely data-driven approach using only the intrinsic spatiotemporal heterogeneity in the time series itself. However, other auxiliary spatiotemporal features could also be utilized if available in the dataset, such as POIs, weather patterns, or urban structural characteristics. These supplemental contextual attributes could be encoded into the query representation to provide further instruction for meta-parameter learning. Therefore, the method is flexible and can leverage diverse contextual features to learn improved meta-parameters tailored to various application domains.

\subsection{Heterogeneity-Informed Spatiotemporal Meta-Network}

Leveraging our proposed techniques, we implement an end-to-end model called Heterogeneity-Informed Spatiotemporal Meta-Network (HimNet) for accurate spatiotemporal forecasting.

Taking inspiration from GCNs, researchers have developed graph convolutional recurrent networks~\cite{DCRNN,GTS,CCRNN} which leverage both graph convolutions and recurrent transformations. By integrating graph convolutions within recurrent cells, these models can effectively capture temporal dynamics while also accounting for the spatial relations in the graph. Therefore, our proposed HimNet architecture uses variants of the widely adopted Graph Convolutional Recurrent Unit (GCRU)~\cite{AGCRN,MegaCRN} as the fundamental building block. GCRU is formulated by:
\begin{equation}
\begin{aligned}
    r_t&=\sigma(\Theta_r\star_{\mathcal{G}}[X_t, H_{t-1}]+b_r)\\
    u_t&=\sigma(\Theta_u\star_{\mathcal{G}}[X_t, H_{t-1}]+b_u)\\
    c_t&=\mathrm{tanh}(\Theta_c\star_{\mathcal{G}}[X_t, (r_t\odot H_{t-1})]+b_c)\\
    H_t&=u_t\odot H_{t-1}+(1-u_t)\odot c_t,
\end{aligned}
\end{equation}
where $X_t\in\mathbb{R}^{N}$ and $H_t\in\mathbb{R}^{N\times h}$ denote the input and output at timestep $t$. $h$ is the hidden size. $r_t$ and $u_t$ are the reset and update gates. $\Theta_r, \Theta_u, \Theta_c$ are parameters for the corresponding filters in the graph convolution operation $\star_{\mathcal{G}}$, which is defined as:
\begin{equation}
    Z=U\star_{\mathcal{G}}=\sum_{k=0}^K\tilde{A}UW_k,
\end{equation}
where $U\in\mathbb{R}^{N\times C}$ is the input, $Z\in\mathbb{R}^{N\times h}$ is the output, and $C$ denotes the input channels. $\tilde{A}$ represents the topology of graph $\mathcal G$, and $W\in\mathbb{R}^{K\times C\times h}$ is the kernel parameter $\Theta$.

By applying temporal meta-parameter learning to GCRU, we propose a \textbf{temporal meta-encoder}. Referring to Equation~\ref{eq: TMP}, the generated temporal meta-parameters in this encoder take the form $W_t\in\mathbb{R}^{B\times K\times C\times h}$. Similarly, we build a \textbf{spatial meta-encoder} via spatial meta-parameter learning, with generated spatial meta-parameters $W_s\in\mathbb{R}^{N\times K\times C\times h}$ by applying Equation~\ref{eq: SMP}. For the graph $\mathcal G$ used in the encoders, rather than relying on a static pre-defined adjacency matrix as in prior works~\cite{STGCN,DCRNN,ASTGCN}, we follow the adaptive graph learning design~\cite{GWNet,MTGNN}. Specifically, we leverage our learned spatial embeddings $E_s$ to dynamically generate the adjacency matrix $\tilde{A}\in\mathbb{R}^{N\times N}$ as:
\begin{equation}
    \tilde A=\mathrm{Softmax}(\mathrm{ReLU}(E_s\cdot E_s^\top)).
\end{equation}
In short, given an input $X^b\in\mathbb{R}^{B\times T\times N}$, the two encoders operate in parallel to encode it into two hidden representations based on their respective meta-parameters. These representations are then summed to yield the final combined hidden encoding $H\in\mathbb{R}^{B\times N\times h}$.

To decode the latent representation $H$ into predictions, we propose a \textbf{spatiotemporal meta-decoder} that leverages ST-mixed meta-parameters. Specifically, we first generate the corresponding spatiotemporal embedding $E_{st}\in\mathbb{R}^{B\times N\times d_{st}}$ from $H$:
\begin{equation}
    E_{st}=H\cdot W_E+b_E,
\end{equation}
where $W_E\in\mathbb{R}^{h\times d_{st}}$ and $b_E\in\mathbb{R}^{d_{st}}$ are the linear projection parameters. According to Equation~\ref{eq: STMP}, we then produce the ST-mixed meta-parameters for GCRU as $W_{st}\in\mathbb{R}^{B\times N\times K\times C\times h}$ using query $E_{st}$. Moreover, inspired by~\cite{MegaCRN}, we apply a time-varying adaptive graph $\mathcal G_{st}$ inferred from $E_{st}$:
\begin{equation}
    \tilde A_{st}=\mathrm{Softmax}(\mathrm{ReLU}(E_{st}\cdot E_{st}^\top)),
\end{equation}
where $\tilde{A}_{st}\in\mathbb{R}^{B\times N\times N}$. The decoder takes the initial hidden state from $H$ and iteratively generates predictions $\hat X$ for each future timestep by applying the GCRU cell parameterized by $W_{st}$.

Given ground truth $X_{t+1:t+T'}$, we mainly adopt the Mean Absolute Error (MAE) loss as our training objective to optimize multi-step predictions jointly. The loss function for HimNet's multi-step spatiotemporal time series forecasting can be formulated as:
\begin{equation}
    \mathcal L=MAE(\hat X_{t+1:t+T'}, X_{t+1:t+T'})=\frac{1}{T'N}\sum_{i=1}^{T'}\sum_{j=1}^N |\hat X_{i,j}-X_{i,j}|.
\end{equation}

\begin{table}[b]
\small
\centering
\caption{Summary of datasets.}
\label{table: datasets}
\renewcommand\arraystretch{1.1}
\begin{tabular}{lccc}
\toprule
\textbf{Dataset} & \textbf{\#Sensors} & \textbf{\#Timesteps} & \textbf{Time Range} \\ 
\midrule
METRLA          & 207                    & 34,272               & 03/2012 - 06/2012   \\
PEMSBAY         & 325                    & 52,116               & 01/2017 - 06/2017   \\
PEMS04           & 307                    & 16,992               & 01/2018 - 02/2018   \\
PEMS07           & 883                    & 28,224               & 05/2017 - 08/2017   \\
PEMS08           & 170                    & 17,856               & 07/2016 - 08/2016   \\
\bottomrule
\end{tabular}
\end{table}

\begin{table*}[t]
\centering
\caption{Performance on METRLA, PEMSBAY, PEMS03, PEMS04, PEMS07, and PEMS08 datasets.}
\label{table: perf}
\renewcommand\arraystretch{1.1}
\resizebox{\linewidth}{!}{%
\begin{tabular}{ccc|ccccccccccccc}
\toprule
\multicolumn{2}{c}{Dataset}                                                                          & Metric & HI      & GRU     & STGCN   & DCRNN   & GWNet              & AGCRN   & GTS     & STNorm  & STID    & ST-WA   & PDFormer & MegaCRN & \textbf{HimNet}             \\
\hline
\multirow{9}{*}{\rotatebox{90}{METRLA}}  & \multirow{3}{*}{\begin{tabular}[c]{@{}c@{}}Step 3\\ 15 min\end{tabular}}  & MAE    & 6.80    & 3.07       & 2.75    & 2.67    & 2.69               & 2.85    & 2.75    & 2.81    & 2.82    & 2.89    & 2.83     & 2.65    & \textbf{2.60}    \\
                         &                                                                           & RMSE   & 14.21   & 6.09        & 5.29    & 5.16    & 5.15               & 5.53    & 5.27    & 5.57    & 5.53    & 5.62    & 5.45     & 5.08    & \textbf{5.02}    \\
                         &                                                                           & MAPE   & 16.72\% & 8.14\%    & 7.10\%  & 6.86\%  & 6.99\%           & 7.63\%  & 7.12\%  & 7.40\%  & 7.75\%  & 7.66\%  & 7.77\%   & 6.73\%  & \textbf{6.70\%}  \\
                         & \multirow{3}{*}{\begin{tabular}[c]{@{}c@{}}Step 6\\ 30 min\end{tabular}}  & MAE    & 6.80    & 3.77       & 3.15    & 3.12    & 3.08                & 3.20    & 3.14    & 3.18    & 3.19    & 3.25    & 3.20     & 3.04    & \textbf{2.95}    \\
                         &                                                                           & RMSE   & 14.21   & 7.69        & 6.35    & 6.27    & 6.20                & 6.52    & 6.33    & 6.59    & 6.57    & 6.61    & 6.46     & 6.18    & \textbf{6.06}    \\
                         &                                                                           & MAPE   & 16.72\% & 10.71\%  & 8.62\%  & 8.42\%  & 8.47\%            & 9.00\%  & 8.62\%  & 8.47\%  & 9.39\%  & 9.22\%  & 9.19\%   & 8.22\%  & \textbf{8.11\%}  \\
                         & \multirow{3}{*}{\begin{tabular}[c]{@{}c@{}}Step 12\\ 60 min\end{tabular}} & MAE    & 6.80    & 4.88        & 3.60    & 3.54    & 3.51               & 3.59    & 3.59    & 3.57    & 3.55    & 3.68    & 3.62     & 3.51    & \textbf{3.37}    \\
                         &                                                                           & RMSE   & 14.21   & 9.75       & 7.43    & 7.47    & 7.28       & 7.45    & 7.44    & 7.51    & 7.55    & 7.59    & 7.47     & 7.39    & \textbf{7.22}       \\
                         &                                                                           & MAPE   & 16.71\% & 14.91\%  & 10.35\% & 10.32\% & 9.96\%   & 10.47\% & 10.25\% & 10.24\% & 10.95\% & 10.78\% & 10.91\%  & 10.01\% & \textbf{9.79\%}     \\
\midrule
\multirow{9}{*}{\rotatebox{90}{PEMSBAY}} & \multirow{3}{*}{\begin{tabular}[c]{@{}c@{}}Step 3\\ 15 min\end{tabular}}  & MAE    & 3.05    & 1.44       & 1.36    & 1.31    & 1.30                & 1.35    & 1.37    & 1.33    & 1.31    & 1.37    & 1.32     & 1.28    & \textbf{1.27}    \\
                         &                                                                           & RMSE   & 7.03    & 3.15        & 2.88    & 2.76    & 2.73                & 2.88    & 2.92    & 2.82    & 2.79    & 2.88    & 2.83     & 2.71    & \textbf{2.68}    \\
                         &                                                                           & MAPE   & 6.85\%  & 3.01\%    & 2.86\%  & 2.73\%  & 2.71\%            & 2.91\%  & 2.85\%  & 2.76\%  & 2.78\%  & 2.86\%  & 2.78\%   & 2.67\%  & \textbf{2.64\%}  \\
                         & \multirow{3}{*}{\begin{tabular}[c]{@{}c@{}}Step 6\\ 30 min\end{tabular}}  & MAE    & 3.05    & 1.97        & 1.70    & 1.65    & 1.63                & 1.67    & 1.72    & 1.65    & 1.64    & 1.70    & 1.64     & 1.60    & \textbf{1.57}    \\
                         &                                                                           & RMSE   & 7.03    & 4.60        & 3.84    & 3.75    & 3.73               & 3.82    & 3.86    & 3.77    & 3.73    & 3.81    & 3.79     & 3.69    & \textbf{3.60}    \\
                         &                                                                           & MAPE   & 6.84\%  & 4.45\%    & 3.79\%  & 3.71\%  & 3.73\%            & 3.81\%  & 3.88\%  & 3.66\%  & 3.73\%  & 3.81\%  & 3.71\%   & 3.60\%  & \textbf{3.52\%}  \\
                         & \multirow{3}{*}{\begin{tabular}[c]{@{}c@{}}Step 12\\ 60 min\end{tabular}} & MAE    & 3.05    & 2.70        & 2.02    & 1.97    & 1.99                & 1.94    & 2.06    & 1.92    & 1.91    & 2.00    & 1.91     & 1.90    & \textbf{1.84}    \\
                         &                                                                           & RMSE   & 7.01    & 6.28        & 4.63    & 4.60    & 4.60                & 4.50    & 4.60    & 4.45    & 4.42    & 4.52    & 4.43     & 4.49    & \textbf{4.32}    \\
                         &                                                                           & MAPE   & 6.83\%  & 6.72\%    & 4.72\%  & 4.68\%  & 4.71\%            & 4.55\%  & 4.88\%  & 4.46\%  & 4.55\%  & 4.63\%  & 4.51\%   & 4.53\%  & \textbf{4.33\%}  \\
\midrule
\multirow{3}{*}{\rotatebox{90}{PEMS03}}  & \multirow{3}{*}{Average}                                                  & MAE    & 32.62   & 19.80     & 15.83   & 15.54   & 14.59     & 15.24   & 15.41   & 15.32   & 15.33   & 15.28   & 14.94    & \textbf{14.57}   & 15.11 \\
                         &                                                                           & RMSE   & 49.89   & 32.75      & 27.51   & 27.18   & \textbf{25.24}     & 26.65   & 26.15   & 25.93   & 27.40   & 26.62   & 25.39    & 25.87   & 26.56 \\
                         &                                                                           & MAPE   & 30.60\% & 18.84\%  & 16.13\% & 15.62\%   & 15.52\%  & 15.89\% & 15.39\% & \textbf{14.37\%} & 16.40\% & 15.17\% & 15.82\%  & 14.75\% & 15.49\% \\
\midrule
\multirow{3}{*}{\rotatebox{90}{PEMS04}}  & \multirow{3}{*}{Average}                                                  & MAE    & 42.35   & 25.55      & 19.57   & 19.63   & 18.53              & 19.38   & 20.96   & 18.96   & 18.38   & 19.06   & 18.36    & 18.72   & \textbf{18.14}   \\
                         &                                                                           & RMSE   & 61.66   & 39.71      & 31.38   & 31.26   & 29.92              & 31.25   & 32.95   & 30.98   & 29.95   & 31.02   & 30.03    & 30.53   & \textbf{29.88}   \\
                         &                                                                           & MAPE   & 29.92\% & 17.35\%  & 13.44\% & 13.59\% & 12.89\%          & 13.40\% & 14.66\% & 12.69\% & 12.04\% & 12.52\% & 12.00\%  & 12.77\% & \textbf{12.00\%} \\
\midrule
\multirow{3}{*}{\rotatebox{90}{PEMS07}}  & \multirow{3}{*}{Average}                                                  & MAE    & 49.29   & 26.74      & 21.74   & 21.16   & 20.47              & 20.57   & 22.15   & 20.50   & 19.61   & 20.74   & 19.97    & 19.83   & \textbf{19.21}   \\
                         &                                                                           & RMSE   & 71.34   & 42.78      & 35.27   & 34.14   & 33.47              & 34.40   & 35.10   & 34.66   & 32.79   & 34.05   & 32.95    & 32.91   & \textbf{32.75}   \\
                         &                                                                           & MAPE   & 22.75\% & 11.58\%  & 9.24\%  & 9.02\%  & 8.61\%            & 8.74\%  & 9.38\%  & 8.75\%  & 8.30\%  & 8.77\%  & 8.55\%   & 8.36\%  & \textbf{8.03\%}  \\
\midrule
\multirow{3}{*}{\rotatebox{90}{PEMS08}}  & \multirow{3}{*}{Average}                                                  & MAE    & 34.66   & 19.36      & 16.08   & 15.22   & 14.40              & 15.32   & 16.49   & 15.41   & 14.21   & 15.41   & 13.58    & 14.75   & \textbf{13.57}   \\
                         &                                                                           & RMSE   & 50.45   & 31.20      & 25.39   & 24.17   & 23.39              & 24.41   & 26.08   & 24.77   & 23.28   & 24.62   & 23.41    & 23.73   & \textbf{23.22}   \\
                         &                                                                           & MAPE   & 21.63\% & 12.43\%  & 10.60\% & 10.21\% & 9.21\%           & 10.03\% & 10.54\% & 9.76\%  & 9.27\%  & 9.94\%  & 9.05\%   & 9.48\%  & \textbf{8.98\%} \\
\bottomrule
\end{tabular}%
}
\end{table*}

\section{Experiment}

\subsection{Experimental Setup}
\noindent\textbf{Datasets.} Our proposed model is evaluated on the \textit{\textbf{five most commonly used}} spatiotemporal forecasting benchmarks. The METRLA dataset and PEMSBAY dataset~\cite{DCRNN} contain traffic speed data recorded by traffic sensors in Los Angeles and the Bay Area, respectively. The PEMS04, PEMS07, and PEMS08 datasets~\cite{STSGCN} are three traffic flow datasets collected from the California Transportation Performance Management System (PeMS). The raw data has a fine temporal granularity of 5 minutes between consecutive timesteps. Therefore, $N_d$=288 and $N_w$=7 in these benchmarks. Additional details regarding the five benchmarks can be found in Table~\ref{table: datasets}. As part of data preprocessing, we perform Z-score normalization on the raw inputs to rescale the data to have zero mean and unit variance. To ensure a fair comparison to previous methods, we adopt the commonly used dataset divisions from prior works. For the METRLA and PEMSBAY datasets, we use 70\% of the data for training, 10\% for validation, and the remaining 20\% for testing~\cite{DCRNN,GWNet}. In the case of the PEMS04, PEMS07, and PEMS08 datasets, we divide them into 60\%, 20\%, and 20\%, respectively~\cite{ASTGCN,STAEformer}. 
In this study, we ensured \textbf{the ethical use} of these five publicly available datasets.

\noindent\textbf{Settings.} We performed hyper-parameter search on each dataset. The encoder and decoder each contain one layer $L$=1 with a hidden dimension $h$=64 for the first four datasets, and $h$=96 for the PEMS08 dataset. 
For the embeddings, the dimensions of the temporal embedding $d_{t}$, the spatial embedding $d_{s}$, and the spatiotemporal embedding $d_{st}$ are all set to 16 for the first four datasets. And for PEMS08, the dimensions are 12, 14, and 10, respectively.
The input and prediction horizons are both set to 1 hour ($T$=$T'$=12 timesteps). We use the Adam optimizer with an initial learning rate of 0.001 and decay over training, and the batch size $B$ is 16. The maximum training epochs is 200, with an early stop patience set to 20. For the loss function, we employ MAE loss for METRLA and PEMSBAY. For PEMS04, PEMS07, and PEMS08, we further employ Huber loss~\cite{huberloss}, a variant of MAE that is more robust. Model performance is evaluated using MAE, Root Mean Square Error (RMSE), and Mean Absolute Percentage Error (MAPE). All following experiments are conducted on NVIDIA GeForce RTX 3090 GPUs.

\noindent\textbf{Baselines.} We compare our HimNet to several widely adopted baselines. HI~\cite{HI} is a standard statistical technique. GRU~\cite{GRU} is a widely adopted univariate time series forecasting method. As for spatiotemporal forecasting, we select several typical models including STGCN~\cite{STGCN}, DCRNN~\cite{DCRNN}, Graph WaveNet~\cite{GWNet}, AGCRN~\cite{AGCRN}, GTS~\cite{GTS}, STNorm~\cite{STNorm}, STID~\cite{STID}, ST-WA~\cite{STWA}, PDFormer~\cite{PDFormer} and MegaCRN~\cite{MegaCRN}. Among them, DCRNN, GTS, and MegaCRN also apply GCRU-based architectures. AGCRN and ST-WA are meta-parameter learning methods closely related to our approach.

\subsection{Performance Evaluation}
The comparison results for the performance of spatiotemporal forecasting are given in Table~\ref{table: perf}. For the METRLA and PEMSBAY datasets, we report performance on 3, 6, and 12 timesteps (15, 30, and 60 minutes). Consistent with previous works, we report the average performance over all 12 predicted timesteps for the PEMS04, PEMS07, and PEMS08 datasets. From the results, we obtain the following observations: (1) Our proposed model, HimNet, significantly outperforms all baselines across all metrics and datasets, demonstrating the effectiveness of our method. By learning meta-parameters based on spatiotemporal heterogeneity, our approach is better able to adapt to diverse spatiotemporal contexts for accurate forecasting. (2) Compared with other GCRU-based models DCRNN, GTS, and MegaCRN, HimNet achieves better performance, highlighting the superiority of our meta-parameter learning scheme. (3) While AGCRN learns parameters only for each spatial node, our method further considers the temporal dimension and utilizes spatiotemporal heterogeneity. ST-WA learns specific parameters for each location and timestep based only on the knowledge learned from the current input. However, our method fully captures and leverages the global spatiotemporal heterogeneity, providing important guidance for more precise local parameter learning. As a result, our HimNet demonstrates better forecasting performance than others.

\subsection{Ablation Study}
In this section, we conduct ablation studies to validate key components of our proposed approach.
We designed the following variants: (1) w/o $E_t$: replaces the temporal embedding with a static matrix of all ones, thus removing temporal heterogeneity modeling. (2) w/o $E_s$: replaces the spatial embedding with a static matrix of all ones, thus removing spatial heterogeneity modeling. (3) w/o $E_{st}$: replaces the spatiotemporal embedding with a static matrix of all ones. (4) w/o TMP: removes temporal meta-parameters by downgrading to randomly initialized parameters. (5) w/o SMP: removes spatial meta-parameters in the same way. (6) w/o STMP: removes ST-mixed meta-parameters in the same way. As shown in Table~\ref{table: ablation}, using identical queries decreases forecasting performance, demonstrating the effectiveness of our heterogeneity-informed approach. Moreover, removing meta-parameters leads to more substantial performance degradation, highlighting the value of applying the spatiotemporal meta-parameters modulated by our method. All these validate that HimNet is complete and indivisible to achieve superior spatiotemporal forecasting performance.

\begin{table}[h]
\small
\centering
\caption{Average MAE of the ablated variants of HimNet.}
\label{table: ablation}
\renewcommand\arraystretch{1.1}
\begin{tabular}{l|ccccc}
\toprule
Model         & METRLA        & PEMSBAY       & PEMS04         & PEMS07         & PEMS08         \\
\hline
w/o $E_t$        & 2.94          & 1.53          & 18.35          & 22.00          & 14.14          \\
w/o $E_s$        & 3.49          & 1.74          & 21.30          & 19.26          & 14.79          \\
w/o $E_{st}$       & 3.07          & 1.55          & 18.55           & 19.77           & 13.61           \\
\hline
w/o TMP       & 2.94          & 1.54          & 18.65          & 19.58          & 14.44          \\
w/o SMP       & 3.53          & 1.75          & 21.41          & 22.26          & 14.07          \\
w/o STMP      & 3.01          & 1.57          & 18.65          & 19.86          & 13.72          \\
\hline
\textbf{HimNet} & \textbf{2.92} & \textbf{1.51} & \textbf{18.14} & \textbf{19.21} & \textbf{13.57} \\
\bottomrule
\end{tabular}%
\end{table}

\subsection{Hyper-Parameter Study}
\begin{figure}[h]
    \centering
    \begin{subfigure}{0.325\linewidth}
        \centering
        \includegraphics[width=\linewidth]{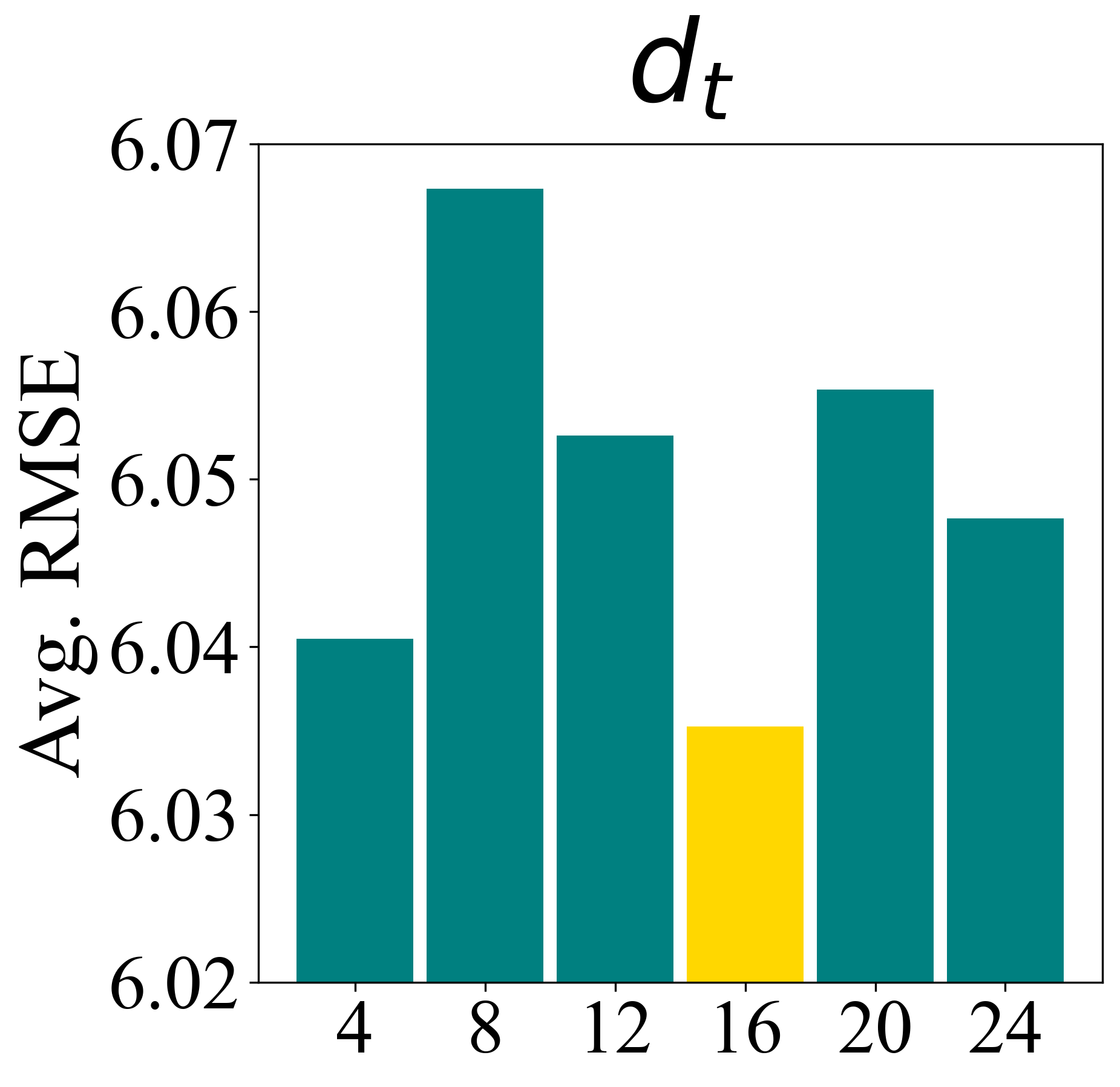}
    \end{subfigure}
    \begin{subfigure}{0.325\linewidth}
        \centering
        \includegraphics[width=\linewidth]{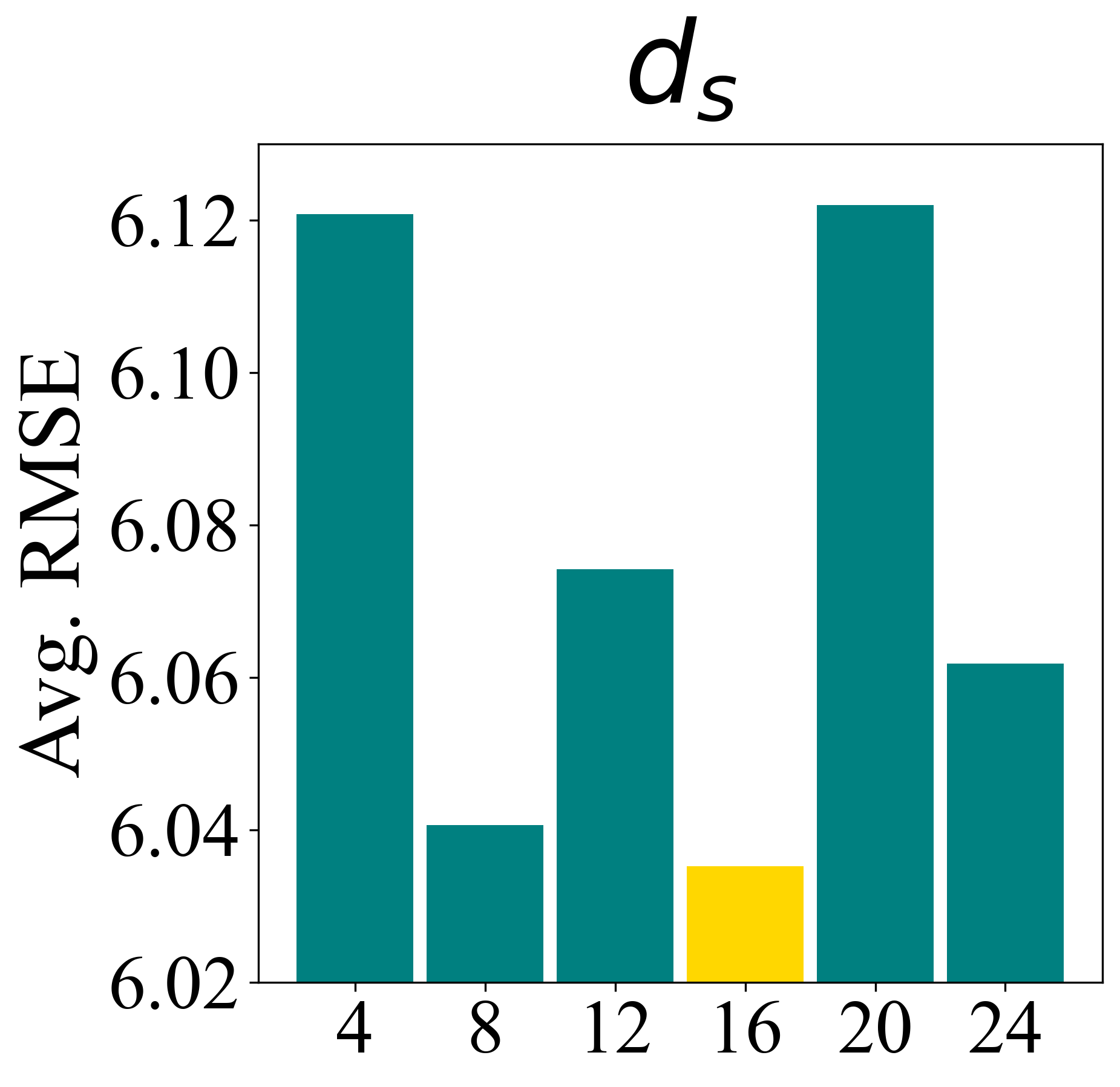}
    \end{subfigure}
    \begin{subfigure}{0.325\linewidth}
        \centering
        \includegraphics[width=\linewidth]{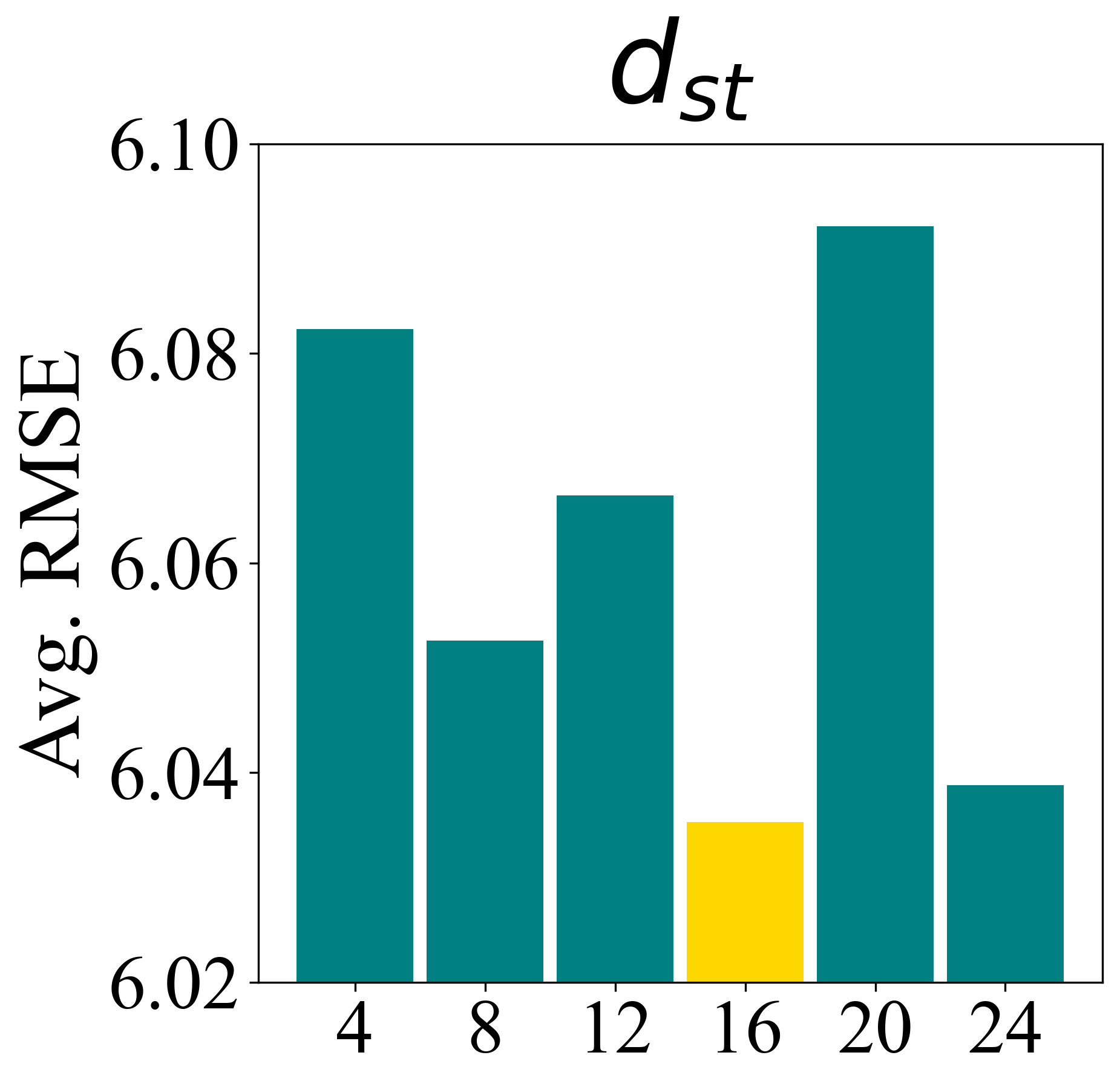}
    \end{subfigure}
    \caption{RMSE w.r.t. embedding dimensions on METRLA.}
    \label{fig: param_study}
\end{figure}

We conduct experiments on our model's sensitivity to key hyper-parameters, including the dimension of the temporal embedding $d_{t}$, the spatial embedding $d_{s}$, and the spatiotemporal embedding $d_{st}$. Figure~\ref{fig: param_study} plots the average RMSE of predictions on the METRLA dataset when varying each embedding dimension from 4 to 24. Interestingly, we find that our model is fairly robust to changes in these hyper-parameters. Across the ranges tested, the RMSE difference between the best and worst-performing configurations is only approximately 0.1. An embedding size of 16 generally provides good results with a tractable parameter size. Still, some trends emerge where decreasing a dimension too much begins to exhibit signs of under-fitting, while larger dimensions will substantially increase parameter size but without notable improvement in performance.

\begin{table}[b]
\centering
\caption{Efficiency comparison on METRLA dataset.}
\label{table: efficiency}
\renewcommand\arraystretch{1.1}
\resizebox{\linewidth}{!}{%
\begin{tabular}{lcccc}
\toprule
Model & \#Params & Time / Batch & Time / Epoch & Mem Usage \\
\midrule
STID~\cite{STID} & 118K & 8ms & 12s & 1420MB\\
\midrule
STGCN~\cite{STGCN} & 246K & 23ms & 34s & 1650MB\\
GWNet~\cite{GWNet} & 309K & 40ms & 60s & 1994MB\\
STNorm~\cite{STNorm} & 224K & 39ms & 59s & 1818MB\\
\midrule
DCRNN~\cite{DCRNN} & 372K & 189ms & 284s & 2134MB\\
AGCRN~\cite{AGCRN} & 752K & 54ms & 82s & 2492MB\\
GTS~\cite{GTS} & 38.5M & 114ms & 171s & 4096MB\\
MegaCRN~\cite{MegaCRN} & 389K & 89ms & 134s & 1962MB\\
\midrule
ST-WA~\cite{STWA} & 375K & 135ms & 203s & 2668MB\\
PDFormer~\cite{PDFormer} & 531K & 173ms & 260s & 6938MB\\
\midrule
HimNet-${\Theta'}$ & 10.9B & N/A & N/A & N/A\\
\textbf{HimNet} & 1251K & 97ms & 144s & 6056MB\\
\bottomrule
\end{tabular}%
}
\end{table}

\subsection{Efficiency Study}
We evaluate efficiency by comparing HimNet with 10 spatiotemporal baseline models. Table~\ref{table: efficiency} reports the number of parameters, per-batch runtime, per-epoch runtime, and GPU memory usage on METRLA (batch size $B$=16). 

As shown in the table, models using TCNs, such as STGCN, GWNet and STNorm, perform relatively well because of the high computational efficiency of convolution. In comparison, the next four models with GCRU architecture show a notable efficiency gap due to the iterative nature of RNNs that intrinsically presents a disadvantage. Additionally, Transformer-based models like ST-WA and PDFormer are even worse due to their quadratic computational complexity. Among these, STID performs best as its backbone is simply linear layers. We also observe that HimNet's memory usage is relatively high, because our design of the meta-parameter pools will enlarge its parameter size. In summary, our HimNet achieves decent efficiency compared with other GCRU-based models, and outperforms the Transformer-based baselines. 

We further analyze a variant HimNet-${\Theta'}$ that removes meta-parameter pools and directly optimizes the three enlarged parameter spaces. HimNet-${\Theta'}$ contains over 10.9 billion parameters, unavailable on any GPUs we can get. This is consistent with our analysis that directly optimizing the enlarged parameter spaces leads to intractable memory costs, which validates the necessity of our meta-parameter pool design to maintain efficiency while enabling flexible parameterization. 




\begin{figure*}[!t]
    \centering
    \includegraphics[width=\linewidth]{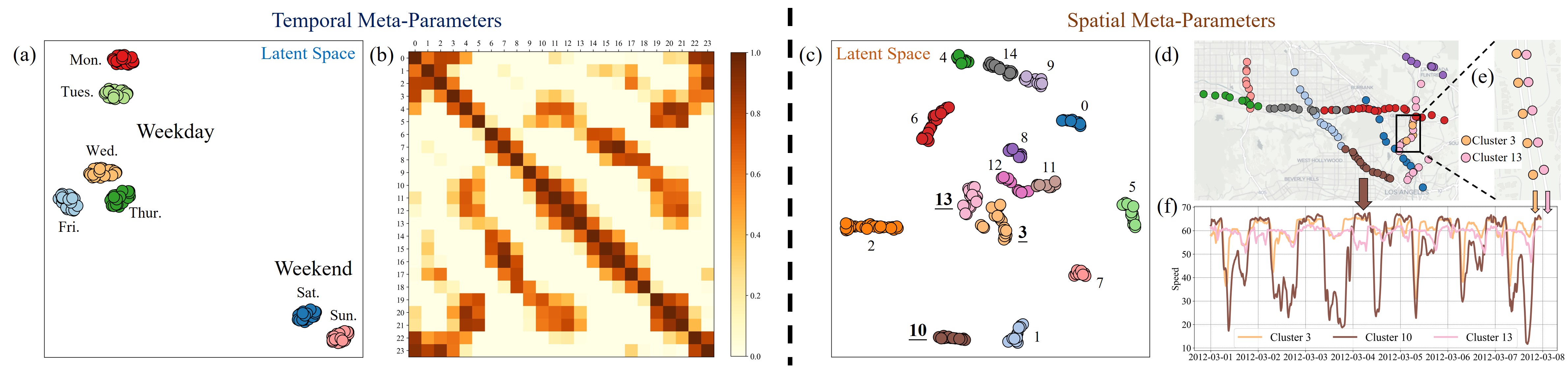}
    \caption{t-SNE visualization of the temporal and spatial meta-parameters. (a): Visualization for each day in a week where every cluster contains 24 points representing each hour's temporal meta-parameter. (b): The cosine similarity matrix of the meta-parameters across 24 hours in a day. (c): Visualization for each sensor's spatial meta-parameter in METRLA dataset. (d): Sensor points in (c) plotted on map, with only one lane of points shown for dual carriageways. (e): Clusters 3 and 13 are located on opposite lanes of the same road. (e): The one-week time series of clusters 3, 10, and 13.}
    \label{fig: S_T_MetaParam}
\end{figure*}

\begin{figure}[!t]
    \centering
    \includegraphics[width=\linewidth]{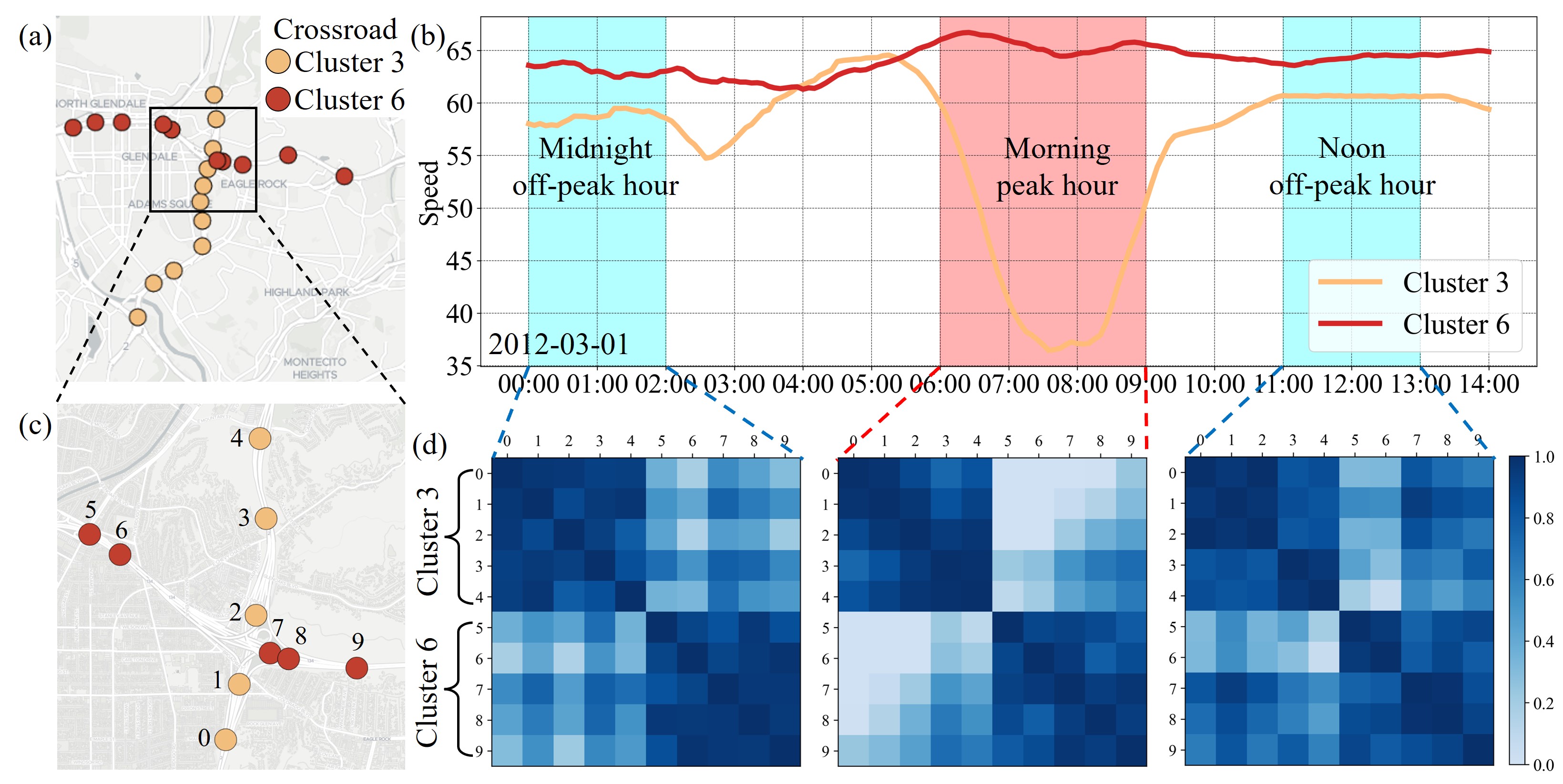}
    \caption{The evolving ST-mixed meta-parameters. (a): Distribution of clusters 3 and 6 around a crossroad area. (b): The traffic speed time series of the two clusters at peak hour and off-peak hour. (c): A closer view of 10 sensors at the crossroad. (d): The cosine similarity of their ST-mixed meta-parameters across three time periods.}
    \label{fig: STM_MetaParam}
\end{figure}

\subsection{Case Study}
In this section, we analyze the learned meta-parameters through visualizations to gain better interpretability and demonstrate how they reflect meaningful circumstances in the real world.

\noindent\textbf{Temporal Meta-Parameters.} Figure~\ref{fig: S_T_MetaParam} provides visualizations on temporal and spatial meta-parameters based on t-SNE~\cite{TSNE} dimensionality reduction. We extract the temporal meta-parameters for each hour across the days of a week. With one day sampled at an hourly granularity, this results in a total of 168 data points embedded in the 2D latent space. As shown in Figure~\ref{fig: S_T_MetaParam}(a), distinct tight clusters form for each individual day, demonstrating our method's ability to differentiate days of the week. Notably, weekdays and weekends are widely separated, underscoring our effectiveness in capturing the temporal heterogeneity between these day types. Figure~\ref{fig: S_T_MetaParam}(b) shows the cosine similarity between the temporal meta-parameters of each hourly interval within a day. Higher similarity is observed between adjacent hours, decreasing with increased temporal distance as expected. Interestingly, the time periods when people travel frequently exhibit high similarity. For example, the periods between 6-9am and 4-7pm show increased similarity, reflecting real-world commuting behaviors. These examples demonstrate that our proposed method successfully identifies and distinguishes the diverse temporal contexts associated with different time periods. 

\begin{figure}[t]
    \centering
    \includegraphics[width=\linewidth]{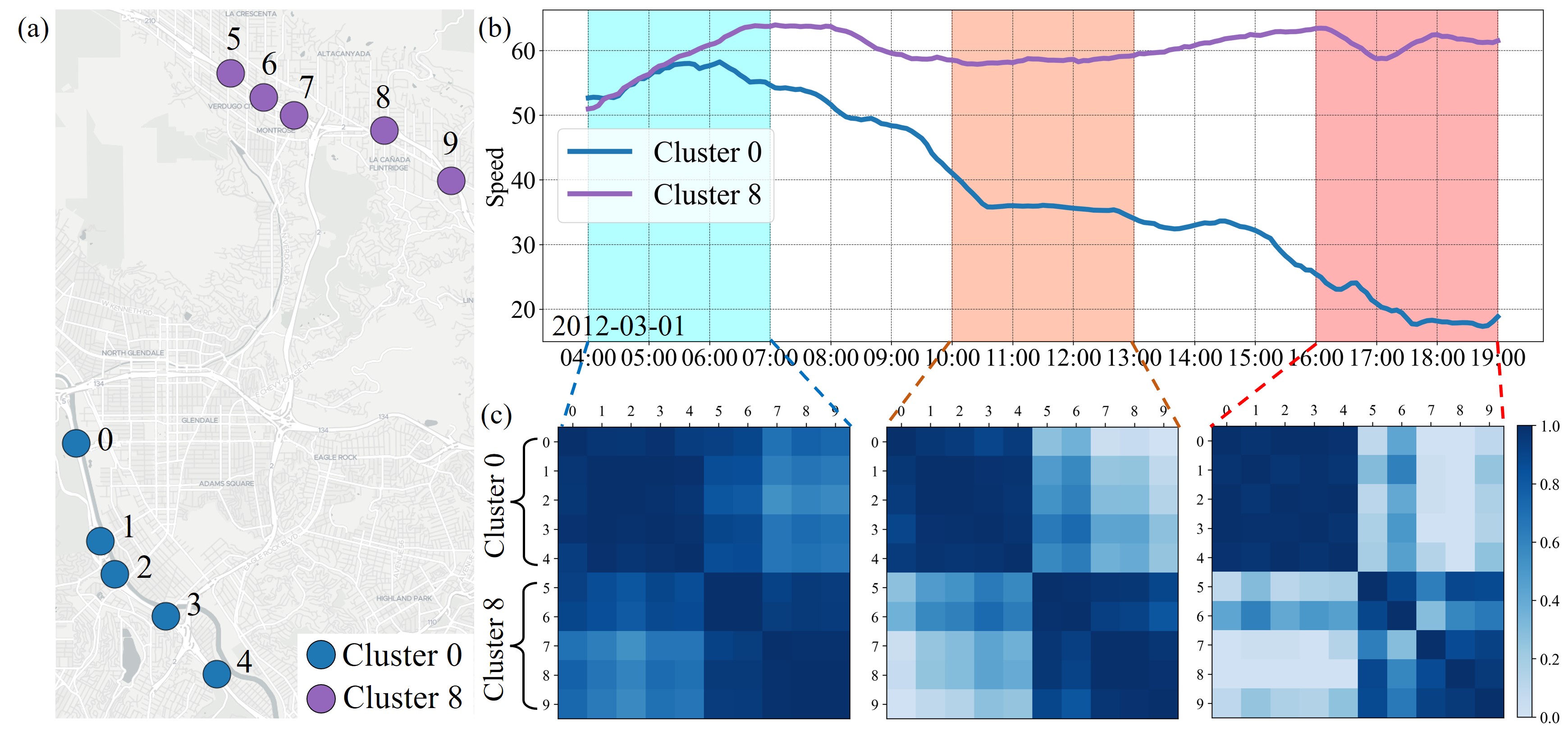}
    \caption{The evolving ST-mixed meta-parameters of clusters 0 and 8. As the time series diverge, they evolve accordingly.}
    \label{fig: case2}
\end{figure}

\noindent\textbf{Spatial Meta-Parameters.} For the spatial meta-parameters, we analyze the METRLA dataset as a case study. It contains 207 sensor locations whose spatial meta-parameters are embedded in 2D latent space as shown in Figure~\ref{fig: S_T_MetaParam}(c). The 207 points naturally form into 15 clusters marked with different colors. To interpret these clusters, we draw them onto a real map of Los Angeles in Figure~\ref{fig: S_T_MetaParam}(d). Remarkably, each cluster precisely aligns with road segments, underscoring our model's strong power to capture real-world urban structural characteristics, even without the help of graph topology. Moreover, opposite lanes of the same road contain points from different clusters. Figure~\ref{fig: S_T_MetaParam}(e) shows clusters 3 and 13 as an example. For visual clarity, we manually select one lane per road to draw Figure~\ref{fig: S_T_MetaParam}(d). This exact matching between learned clusters and real road topology demonstrates our method's ability to explicitly distinguish spatial heterogeneity. Furthermore, we analyze the time series of clusters 3, 10, and 13 specifically. Cluster 10 corresponds to a major downtown road prone to severe congestion. Clusters 3 and 13 are located on the opposite lanes of a minor road. As expected, their traffic speed time series in Figure~\ref{fig: S_T_MetaParam}(f) show that clusters 3 and 13 are closely matched while significantly differing from 10, consistent with our observations on the map and the latent space in Figure~\ref{fig: S_T_MetaParam}(c). This validates that the spatial meta-parameters encode meaningful contexts reflecting spatial heterogeneity.

\noindent\textbf{ST-Mixed Meta-Parameters.} Figure~\ref{fig: STM_MetaParam} provides a case study of the evolving ST-mixed meta-parameters. Clusters 3 and 6 from the crossroad area in Figure~\ref{fig: STM_MetaParam}(a) are analyzed. Their traffic speed time series are shown in Figure~\ref{fig: STM_MetaParam}(b) with three periods highlighted: midnight off-peak, morning peak, and noon off-peak. Ten nearby sensors (5 per cluster) on the crossroad are visualized in Figure~\ref{fig: STM_MetaParam}(c). For their ST-mixed parameters, we draw cosine similarity heatmaps across the three time periods in Figure~\ref{fig: STM_MetaParam}(d). As expected, within-cluster similarity is generally higher than cross-cluster similarity. In detail, during midnight off-peak hours, both clusters exhibit similar traffic speed patterns, reflected by high cross-cluster similarity. However, during the morning peak, cluster 3 shows a sharp drop in speed while cluster 6 maintains a higher level. Their time series diverge, resulting in lower cross-cluster similarity. At noon, the speed patterns in both clusters become closer, leading to an increase in cross-cluster similarity again.
Moreover, in Figure~\ref{fig: case2}, we conduct another case study on clusters 0 and 8. Figure~\ref{fig: case2}(a) shows five sensors each selected from the two clusters. As shown in Figure~\ref{fig: case2}(b), their time series diverge substantially over time. Correspondingly, as illustrated in Figure~\ref{fig: case2}(c), the cross-cluster similarity between the ST-mixed meta-parameters associated with each sensor decreases accordingly over the three selected time periods.
These case studies intuitively explain how our ST-mixed meta-parameters evolve spatially and temporally. Thus, the proposed approach has strong adaptability to various spatiotemporal contexts, with meta-parameters accurately capturing changes between different time periods for each location. Furthermore, this also confirms the power of our method to jointly model spatiotemporal heterogeneity.

\section{Conclusion}
In this study, we proposed a novel \textbf{Heterogeneity-Informed Meta-Parameter Learning} scheme along with the state-of-the-art Heterogeneity-Informed Spatiotemporal Meta-Network (\textbf{HimNet}) for spatiotemporal forecasting. In detail, we captured spatiotemporal heterogeneity by learning spatial and temporal embeddings. A novel meta-parameter learning paradigm was proposed to learn spatiotemporal-specific parameters from meta-parameter pools. Critically, our proposed approach can fully leverage the captured spatiotemporal heterogeneity to inform meta-parameter learning. Extensive experiments on five benchmarks demonstrated HimNet's superior performance. Further visualization analyses on meta-parameters revealed its strong interpretability.

\section*{Acknowledgment}
This work was partially supported by the grants of National Key Research and Development Project (2021YFB1714400) of China, Jilin Provincial International Cooperation Key Laboratory for Super Smart City and Zhujiang Project (2019QN01S744).

\bibliographystyle{ACM-Reference-Format}
\bibliography{reference}

\appendix
\section{Appendix}
\subsection{Notation Table}
For reference, Table~\ref{table: notation} summarizes the key notations and their descriptions in this paper.

\begin{table}[h]
\small
\centering
\caption{Notations and their descriptions.}
\label{table: notation}
\renewcommand\arraystretch{1.1}
\begin{tabular}{l|l}
\toprule
Notation & Description \\
\hline
$B$ & Batch size.\\
$N$ & Number of spatial locations.\\
$T$, $\mathcal T$ & Number of time steps.\\
$X$, $X^b$ / $\hat X$ & Input/output time series.\\
$E_t$, $E_s$, $E_{st}$ & Embedding matrices.\\
$d_{tod}$, $d_{dow}$, $d_{t}$, $d_{s}$, $d_{st}$ & Embedding dimensions.\\
$\Theta$, $\Theta'$, $\Theta_t$, $\Theta_s$, $\Theta_{st}$ & Parameter spaces.\\
$P$, $P_t$, $P_s$, $P_{st}$ & Meta-parameter pools.\\
$\vartheta$, $\vartheta_t$, $\vartheta_s$, $\vartheta_{st}$ & Meta-parameters.\\
$S$ & Parameter size.\\
$H$, $h$ & GCRU hidden encoding and its size.\\
$\tilde A$, $\tilde A_{st}$ & Adaptive graph adjacency matrices.\\
\bottomrule
\end{tabular}%
\end{table}



\subsection{Training Settings}
For reproducibility, we provide details of our training settings in Table~\ref{table: train_setting}. Different values used for METRLA, PEMSBAY, PEMS04, PEMS07, and PEMS08 are separated by slashes. This could also be found in the configuration file of our public code.

\vfill\eject

\begin{table}[h]
\small
\centering
\caption{Detailed training settings.}
\label{table: train_setting}
\renewcommand\arraystretch{1.1}
\begin{tabular}{l|l}
\toprule
Configuration & METRLA/PEMSBAY/PEMS04/PEMS07/PEMS08 \\
\hline
Batch size ($B$) & 16\\
Optimizer & Adam\\
Weight decay & 0.0005/0.0001/0.0001/0.0001/0\\
Epsilon & 0.001\\
Learning rate (LR) & 0.001\\
Scheduler milestones & [30, 40]/[25, 35]/[30, 50]/[40, 60]/[40, 60, 80]\\
Scheduler LR decay & 0.1\\
Early stop patience & 20\\
Max number of epochs & 200\\
Gradient clip & 5\\
\bottomrule
\end{tabular}%
\end{table}

\subsection{Efficiency Report}
\begin{figure}[h]
    \centering
    \includegraphics[width=\linewidth]{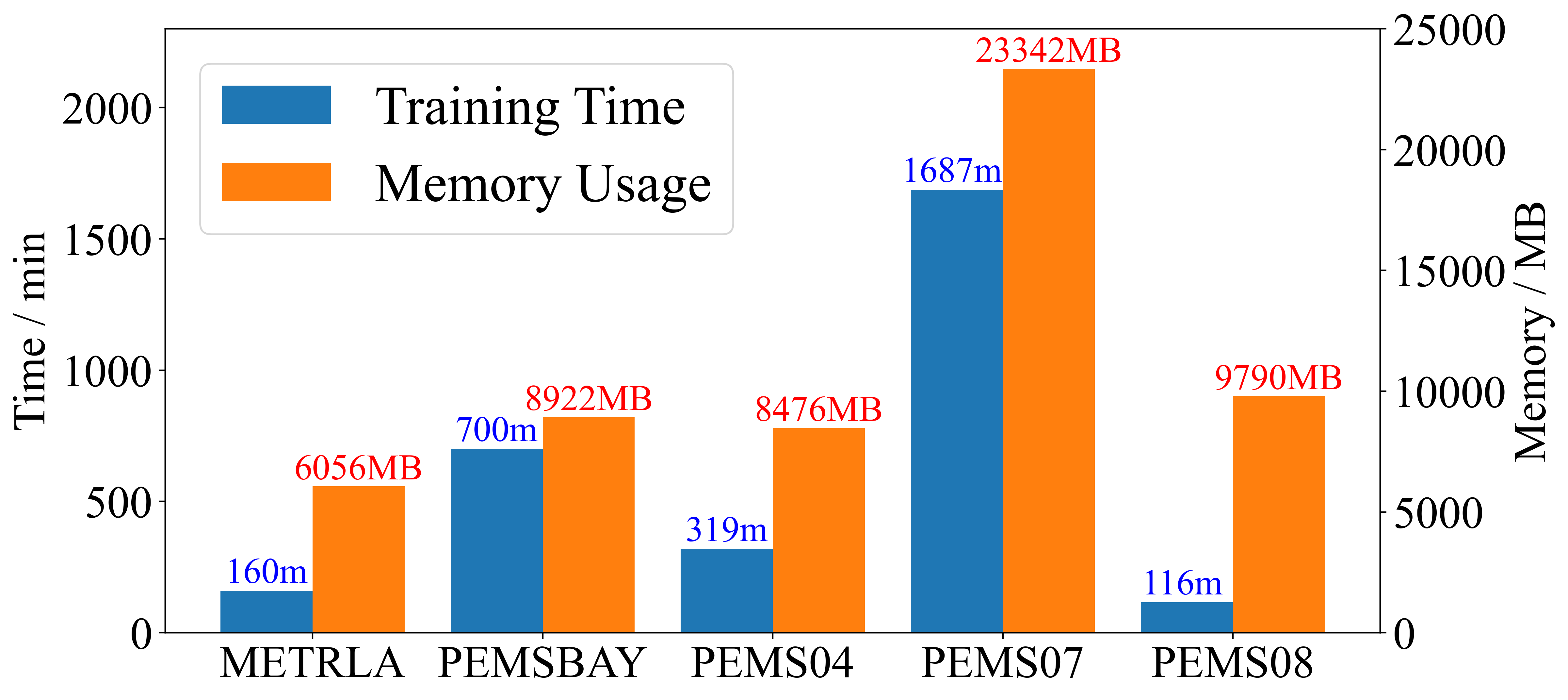}
    \caption{Summary of training time and memory cost.}
    \label{fig: time_mem}
\end{figure}

All the experiments are performed on an Intel(R) Xeon(R) Silver 4310 CPU @ 2.10GHz, 256G RAM computing server, equipped with NVIDIA GeForce RTX 3090 graphics cards. We report the total training time and memory cost in Figure~\ref{fig: time_mem}. Even for the largest dataset PEMS07 (883 sensors), our memory usage is still under 24GB, allowing running all benchmarks on a single RTX 3090 GPU.

\end{document}